\documentclass{article}

\PassOptionsToPackage{numbers, compress}{natbib}
\usepackage[final]{nips_2017}

\usepackage[utf8]{inputenc} 
\usepackage[T1]{fontenc}    
\usepackage{hyperref}       
\usepackage{url}            
\usepackage{booktabs}       
\usepackage{amsfonts}       
\usepackage{nicefrac}       
\usepackage{microtype}      
\usepackage{amsthm}
\usepackage{amsmath}
\usepackage{amsfonts}
\usepackage{bbm}
\usepackage{graphicx}
\usepackage{subfig}
\usepackage[]{algorithm2e}
\usepackage{arydshln}
\usepackage{textcomp}
\usepackage{framed}
\usepackage{color} 

\usepackage{rotating}
\usepackage{arydshln}

\usepackage{titlesec}
\titlespacing*{\section}{0pt}{0pt}{0pt}

\newcommand{\anote}[2]{[{\color{red}{#1}} - {\color{blue} Note: {#2}}]}

\title{A Double Parametric Bootstrap Test for Topic Models}

\author{
  Skyler Seto\thanks{Authors contributed equally.},  Sarah Tan\footnotemark[1], Giles Hooker, Martin T. Wells\\
  Department of Statistical Science\\
  Cornell University\\
  Ithaca, NY 14850 \\
  \texttt{$\{$ss3349, ht395, gjh27, mtw1$\}$@cornell.edu} \\
}

\begin{document}

\maketitle

\begin{abstract}
Non-negative matrix factorization (NMF) is a technique for finding latent representations of data. The method has been applied to corpora to construct topic models. However, NMF has likelihood assumptions which are often violated by real document corpora.  We present a double parametric bootstrap test for evaluating the fit of an NMF-based topic model based on the duality of the KL divergence and Poisson maximum likelihood estimation. The test correctly identifies whether a topic model based on an NMF approach yields reliable results in simulated and real data.
\end{abstract}

\section{Introduction}
Non-negative matrix factorization (NMF) is a technique for decomposing a matrix $X$ with non-negative entries into a low-rank approximation $\hat{X} = WH$ where both $W$ and $H$ have a low rank less than $k$ and contain no negative entries. In early work on NMF , Lee and Seung provide algorithms for computing $W$ and $H$ and demonstrate that the computed factors contain latent representations of face images which their paper notes to have visual interpretation \cite{lee1999learning, lee2001algorithms}. In other work, Xu et al. used NMF to construct topic models \cite{xu2003document}.

Several methods for evaluating topic models have been proposed, such as coherence, perplexity, cosine similarity, topic log odds, etc. \cite{chang2009reading,pauca2004text, wallach2009evaluation}. Less work has been proposed to check if the assumptions behind topic models are valid on the text corpus of interest. 

In this paper, we propose a statistical hypothesis test based on the double parametric bootstrap (DPBS), and duality between minimization of the generalized KL-divergence and maximum likelihood estimation for Poisson random variables for quantifying the fit of a given factorization to a text corpus. This provides a single well-understood probability (p-value) that quantifies the goodness of fit of the topic model to the text corpus. 

Our work is related to the bayesian checking methods proposed by Mimno and Blei for Latent Dirichlet Allocation (LDA) based topic models \cite{mimno2011bayesian}. There, they propose bayesian posterior predictive checks to verify the validity of LDA's statistical independence assumptions. Our approach is analogous, proposing a way to check the statistical distributional assumptions underlying NMF. This can provide NMF topic modeling practitioners more insight into when the topic model does not fit the text corpus, which contributes to the interpretability of the method. 

\section{The Duality between Minimizing KL Divergence and the Poisson Distributional Assumption of NMF}
\label{duality}

In NMF, a matrix $X \in \mathbb{R}^{V\times M}$ of all non-negative entries is decomposed into two non-negative factors $W$ and $H$ that have latent dimensionality $k$, such that $X \approx WH = \hat{X}$. 
The optimal $W$ and $H$ matrices are found by minimizing $D(X \| \hat{X})$, i.e., the distance between $X$ and its low-rank approximation. We illustrate the proposed method with $D(X \|\hat{X}) = -\left(\sum_{i,j} x_{ij}\log\left(\frac{\hat{x}_{ij}}{x_{ij}}\right) - \hat{x}_{ij} + x_{ij}\right)$ where $D(\cdot)$ is the generalized KL-divergence. Our method also extends to other divergences such as Frobenius norm or $\beta$-divergence which we leave for future work.

Prior work demonstrated that NMF has a statistical interpretation as maximum likelihood estimation \cite{cemgil2009bayesian}. To see this, note that if $X_{ij} \sim \text{ Pois}(\lambda_{ij})$ then the conditional likelihood of $X$ given $\Lambda$ is $\log P(X | \Lambda) =\sum_{i,j} x_{ij}\log\left(\lambda_{ij}\right) - \lambda_{ij}  - \log\left(\Gamma(x_{ij}+1)\right)$. Taking $\Lambda=\hat{X}$, this tells us that minimizing generalized KL divergence and maximum likelihood estimation for a Poisson
random variable are equivalent optimization problems. We refer to this as the distributional assumption of NMF. We refer the reader to \cite{brown1986fundamentals} for additional results on the theory of convex analysis and Fenchel dual in general exponential families.


\section{Method: Double Parametric Bootstrap Test for NMF}
\label{bootstrap}

We now provide a statistical hypothesis test for checking the Poisson distributional assumption which evaluates whether the minimum $\hat{X}$ of $D(\cdot)$ is a good estimate. The algorithm determines if the entries of $X$ are drawn from a Poisson distribution using a double parametric bootstrap test. The double parametric bootstrap test was first proposed by Beran in \cite{beran1987prepivoting}; it works well for our purposes because there are no natural pivotal\footnote{Pivots are functions of the data and unknown parameter values whose distribution does not depend on unknown parameters.} test statistics for our hypothesis. Typically the existence of a pivotal statistic is necessary for an exact test procedure.

Classic bootstrap tests based on asymptotically pivotal test statistics generally perform better in finite samples than tests based on asymptotic theory \cite{mackinnon2001improving}. 
However, bootstrap tests do not always perform well in finite samples. Since the true p-value depends on the unknown underlying distribution, while the bootstrap p-value is based on the distribution of the bootstrap statistics (which in turn depends on the bootstrap data generating process), these two distributions will differ whenever the test statistic used is not pivotal and the parameter estimates used in the bootstrap data generating process differs from the true values of the parameter. However, if the test statistic is asymptotically pivotal, the double bootstrap distribution will converge to the true one as the sample size increases. Beran showed in \cite{beran1988prepivoting}, the double bootstrap p-value value will converge to the true p-value at a rate faster than does the asymptotic p-value. Algorithm \ref{alg1} is an application of a  DPBS goodness-of-fit test of  based on an equal-tailed p-value to assess if $X$ satisfies the Poisson distributional assumption.

\begin{algorithm}
 \caption{Double Parametric Bootstrap for Topic Models}
 \begin{framed}

 \KwData{$X$}
 \KwResult{p-value $\rho$}
 Compute $\hat{X}$ for the observed $X$ and let $\ell = \frac{D(X \|\hat{X})}{\sqrt{V\cdot D}}$\;
 Sample $B_1$ bootstrap samples $X^*_1, X^*_2, \dots, X^*_{B_1}\sim \text{ Pois}(\hat{X})$\;
 \For{$i=1:B_1$}{
  Compute $\hat{X}^*_i$ and $\ell^*_i = \frac{D(X^*_i \|\hat{X}^*_i)}{\sqrt{V\cdot D}}$\;
 }
 Compute $\rho^*(\ell) = 2\min\left\{\frac{1}{B_1}\sum_{i=1}^{B_1} \mathbbm{1}\left[\ell^*_i \leq \ell\right], \frac{1}{B_1}\sum_{i=1}^{B_1} \mathbbm{1}\left[\ell^*_i > \ell\right]\right\}$\;
 
 \For{$i=1:B_1$}{
 	Sample $B_2$ bootstrap samples $X^{**}_{i1}, \dots X_{iB_2}^{**} \sim \text{ Pois}(\hat{X}^*_i)$\;
    \For{$j=1:B_2$}{
    	Compute $\hat{X}^{**}_{ij}$ and $\ell^{**}_{ij} = \frac{D(X^{**}_{ij} \|\hat{X}^{**}_{ij})}{\sqrt{V\cdot D}}$\;
    }
    Compute $\rho_{i}^{**}(\ell^*_i) =  2\min\left\{\frac{1}{B_2}\sum_{j=1}^{B_2} \mathbbm{1}\left[\ell^{**}_{ij} \leq \ell^*_i\right], \frac{1}{B_2}\sum_{j=1}^{B_2} \mathbbm{1}\left[\ell^{**}_{ij} > \ell^*_i\right]\right\}$\;
 }
 \Return $\rho =  2\min\left\{\frac{1}{B_1}\sum_{i=1}^{B_1} \mathbbm{1}\left[\rho^* \leq \rho^{**}_i\right], \frac{1}{B_1}\sum_{i=1}^{B_1} \mathbbm{1}\left[\rho^* > \rho^{**}_i\right]\right\}.$
 \end{framed}
 \label{alg1}
\end{algorithm}

An alternative to DPBS is the residual bootstrap. A drawback of
the residual bootstrap for this data is that X is a count matrix that has a heteroscedastic variance structure.  To account for heteroscedasticity in a linear model, Wu in \cite{wu1986jackknife} and Mammen in \cite{mammen1993bootstrap} proposed the wild bootstrap, randomly weighting the residuals.  If the residuals are Gaussian with constant variance, residuals will also have constant variance, however Poisson residuals are heteroscedastic and thus the classical residual bootstrap will perform poorly compared to the double parametric bootstrap. 

\section{Experiments}
\label{res}
In this section, we demonstrate empirical results of DPBS on simulated data from known distributions, and in real-world corpora by studying groupings of the documents. Our implementation of DPBS uses an implementation\footnote{https://gist.github.com/omangin/8801846} of NMF with generalized KL divergence developed by Oliver Mangin.

\vspace{-4.5mm}

\subsection{Simulations}
\label{simulation}
DPBS checks whether the data in $X$ satisfies the distributional assumption of NMF. We present synthetic data experiments evaluating whether the model incorrectly discovers violations in model assumptions when the data is generated from a Poisson distribution. We show that in these cases, the model does not incorrectly detect model assumption violations as to be expected. We further demonstrate empirically that the distributions of p-values is indeed uniform on $[0,1]$.

We simulate three differently sized corpora with $W = 16, 23, 23$ words, $M = 10, 23, 92$ documents, and $K = 5, 10, 10$ topics respectively.  $X \in \mathbb{R}^{10 \times 16}$ is chosen as a small example and is also used in \cite{cemgil2009bayesian}. We generate $W \sim \text{ Gamma}(10, 0.1)$ and $H \sim \text{ Gamma}(1, 100)$  as the Gamma distribution is conjugate to the Poisson and is used as a prior in Bayesian NMF frameworks such as \cite{cemgil2009bayesian} and compute $\hat{X}=WH$. We then generate $X \sim \text{ Pois}(\hat{X})$. In these preliminary experiments we fix $B_1, B_2 = 25$.

To test for uniformity, we generate $100$ different $X$'s for each of the three corpora, and perform a Kolmogorov-Smirnov test (KS test). The null hypothesis of the test asks ``is the sample drawn from the known distribution?'' In this study, the test statistic $T$ and p-value for the KS test are calculated by comparing the $100$ p-values obtained from the $100$ $X$'s and $100$ random samples from a uniform distribution. Results are shown for the KS test in Table \ref{simulated_table}.  In each of the three synthetic corpora, the p-value is not statistically significant at $95\%$ confidence. This confirms the test does not find assumption violations when none exist.


Probability-probability (P-P) plots with error bars of size $\pm T$ are shown in the Appendix in Figures \ref{fig:1016}-\ref{fig:9223}. In all cases the simulated p-values are close to the line $y=x$ and are well-within the error bars $\pm T$. This confirms again that no violations are found.

We also evaluate whether DPBS detects known violations. We sample $25$ $X\in \mathbb{R}^{10\times 16}$ with $k = 5$ from a Zero Inflated Poisson with $p=0.5$, Gamma, and Normal distribution with negative values replaced by zero, and report the average distance $\ell$ and p-value over the $25$ samples in Table \ref{simulated_table}. We compute $\hat{X} = WH$ as before and take all parameter(s) of the distributions to be $\hat{X}$. The reconstruction error $\ell$ is significantly higher when $X$ is not generated from a Poisson, and DPBS finds significant violations for all samples.

\begin{table*}
\centering
    \begin{tabular}{|c | c|}
        \hline
		Size & p-value $\rho$\\
        \hline
        $10 \times 16 $ & 0.3685\\
        \hline
        $23 \times 23 $ & 0.1902\\
        \hline 
        $92 \times 23$ & 0.3687\\\hline
        \end{tabular}
     \quad
     \begin{tabular}{|c | c | c|}
        \hline
		Distribution & p-value $\rho$ & $\ell$\\
        \hline
        Poisson & 0.4416& 2.0277\\
        \hline
        Gamma & 0.0& 859.9227\\
        \hline 
        Normal & 0.0 & 442.5251\\
        \hline
        Zero Inflated Poisson & 0.0& 471.7884\\\hline
        \end{tabular}
    \caption{The table on the left shows results from the KS test for each of the three Poisson-generated synthetic corpora. The reported $T$ and p-value are averaged over twenty iterations as the KS test randomly samples from a uniform distribution.  The table on the right shows the average p-value and $\ell$ from $25$ iterations of the DPBS test when $X$ is drawn from the specified distribution.}
         \label{simulated_table}

\end{table*}

\vspace{-3mm}
\subsection{Use Case: Detecting Group Structure Across Documents}
\label{grouping}
We present a use case of how this method can be used to detect group structure across documents. 
Mimno and Blei studied the New York Times Annotated Corpus\footnote{\url{http://www.ldc.upenn.edu}}, using their proposed bayesian checking test for LDA-based topic models \cite{mimno2011bayesian} to detect variation in word usage within topics by time and perspective (desk that produced the article, e.g. foreign desk). Specifically, they show that for the twenty words that make up the Iraq topic\footnote{\texttt{troops, leaders, military, forces, country, city, security, iraq, iraqi, hussein, baghdad, saddam, shiite, kurds, kurdish, sunni, sadr, iraqis, government, al}}, certain words are more prominent during certain time periods (e.g. kurdish is prominent during the Gulf War but not during the Iraq War), suggesting that the LDA topic model could be improved by taking into account time. As for desk, one hypothesis is that the twenty words of the Iraq topic may by more frequently used in articles from the foreign desk and less in articles from the arts and culture desk. For example, Mimno and Blei found that the word leaders varies more by desks than by time \cite{mimno2011bayesian}.

We study the same set of words to see if the proposed test for NMF topic models also detects temporal and desk structure across documents. The left part of Table \ref{table: grouppvals} describes the results from investigating temporal structure for one desk - the Foreign desk; the right part describes the results of investigating desk structure for one year - 2005. The experimental setup in detailed in the Appendix. Briefly, in the left setting, the four document-term matrices are grouped by time; in the right setting, they are grouped by desk. We operationalize the null hypothesis of no group structure as a study of how our proposed test performs on each $X_1$, $X_2$, $X_3$, $X_4$ separately, as opposed to when all four matrices are combined into one, $X = [X_1$ \textbrokenbar $X_2$ \textbrokenbar $X_3$  \textbrokenbar $X_4]^T$. 


Table \ref{table: grouppvals} lists the p-values obtained from the DPBS test. For both the temporal (left) and desk (right) settings, when the test is performed on the entire document-term matrix $X$, the p-value is 0, hence the likelihood assumption behind NMF is violated. Examining the matrix directly suggests that zero-inflation could be a problem, with $X$ being extremely sparse. However, running the same test with $X$ broken up by time (left) or desk (right) tells us that the likelihood assumption is not violated across all entries of $X$ - in the temporal setting (left) the assumption might only be violated for 2005 and 2006, suggesting that accounting for temporal differences in the topic discourse patterns warrants further study; in the desk setting (right) only the p-value for the $X_3$, the Arts desk matrix, is significant. This suggests that these words are used differently by the Arts/Cultural desk compared to the other three desks, and capturing this grouping in the topic model may improve the topic model.

We note that we condition on only one desk when investigating temporal groupings, and on only one year when investigating desk groupings, to attempt to disentangle temporal effects from desk effects. Results for all desks and all years are in Table \ref{table: grouppvalsall} in the Appendix.

\begin{table}[t!]
\centering
\begin{tabular}{| c c c c|}
\hline
Matrix & Scope & p-value $\rho$ & Rejects\\
\hline
$X$ &  & 0.0 & 10 \\
\hdashline
$X_1$ & 2004 & 0.32 & 2 \\
$X_2$ & 2005& 0.02 & 8\\
$X_3$ & 2006& 0.07 & 8\\
$X_4$ & 2007& 0.76 & 0\\
\hline
\end{tabular}
\begin{tabular}{| c c c c|}
\hline
Matrix & Scope & p-value $\rho$ & Rejects\\
\hline
$X$ & & 0.0 & 10 \\
\hdashline
$X_1$ & Foreign & 0.49 & 1 \\
$X_2$ & Business & 0.06 & 8 \\
$X_3$ &Arts and Culture & 0.0 & 10\\
$X_4$ & National & 0.73 & 0\\
\hline
\end{tabular}
\caption{\textbf{Left}: Testing for temporal structure in New York Times articles from the \textbf{Foreign desk}: p-values for proposed parametric bootstrap test on entire document-term matrix vs. same matrix broken into four sub-matrices by time. \\
\textbf{Right}: Testing for desk structure in New York Times articles for \textbf{January 2005}: p-values for proposed parametric bootstrap test on entire document-term matrix vs. same matrix broken into four sub-matrices by desk.}
\label{table: grouppvals}
\end{table}

\section{Conclusion and Future Work}
\label{conc}
In this work, we present a DPBS test for checking violations to a distributional assumption, namely Poisson, of NMF. We show that the proposed test reliably discovers violations where they exist while controlling falso positives. When distributional assumptions for $X$ are violated, alternative divergence measures such as $\alpha$ or $\beta$-divergence can be considered. These correspond to a Tweedie data generating distributions, as explored by Kameoka in \cite{kameoka2016non}. The Tweedie compound Poisson distribution is a subclass of the exponential dispersion family with a power variance function. We are currently extending our proposed DPBS test for these divergence-distribution pairs. It is our hope that the DPBS procedure can help researchers identify the best divergence measure in NMF for their data. 


\bibliography{interpret_symp}
\bibliographystyle{plain}

\newpage
\section*{Appendix}
\subsection*{PP-Plots}
\begin{figure}[h!]
\minipage{0.32\textwidth}
   \includegraphics[width=\textwidth]{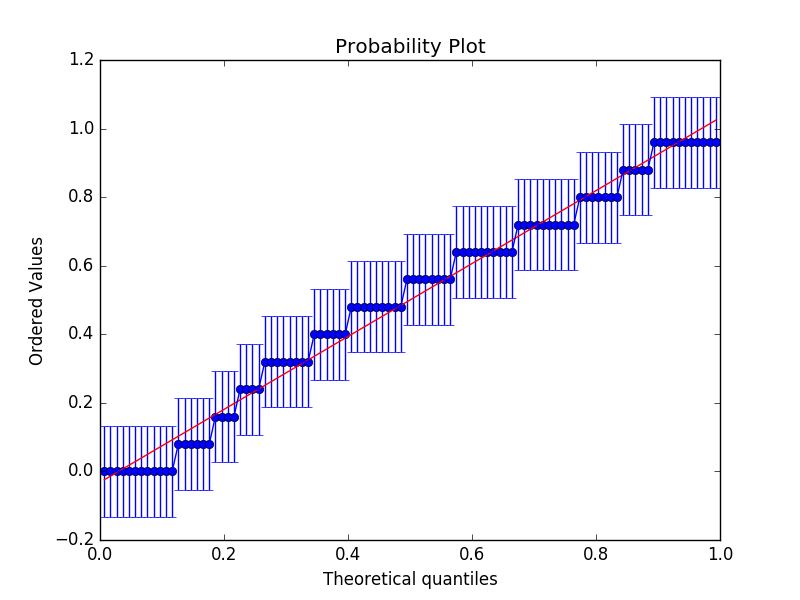}arydshln
  \caption{P-P plot for $W = 10, M = 16, K = 5, T = 0.133$}\label{fig:1016}
\endminipage\hfill
\minipage{0.32\textwidth}
   \includegraphics[width=\textwidth]{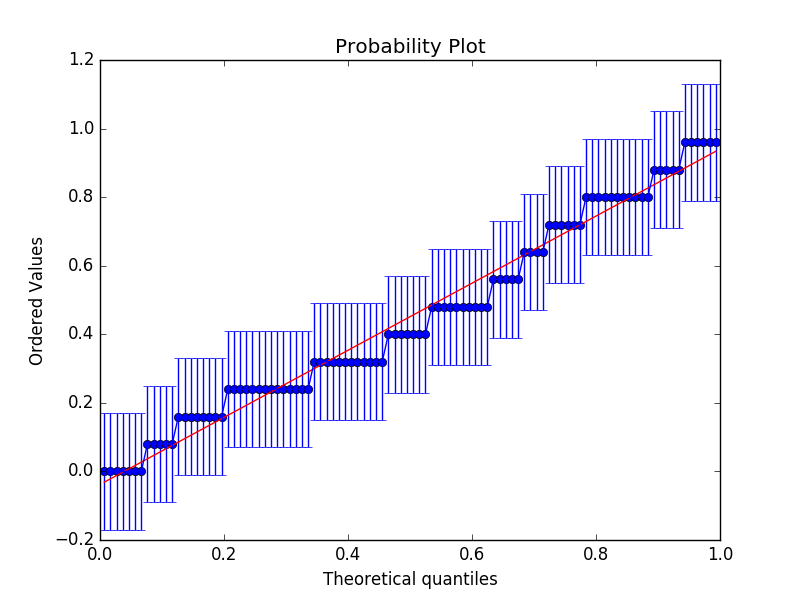}
  \caption{P-P plot for $W = 23, M = 23, K = 10, T = 0.17$}\label{fig:2323}
\endminipage\hfill
\minipage{0.32\textwidth}%
 \includegraphics[width=\textwidth]{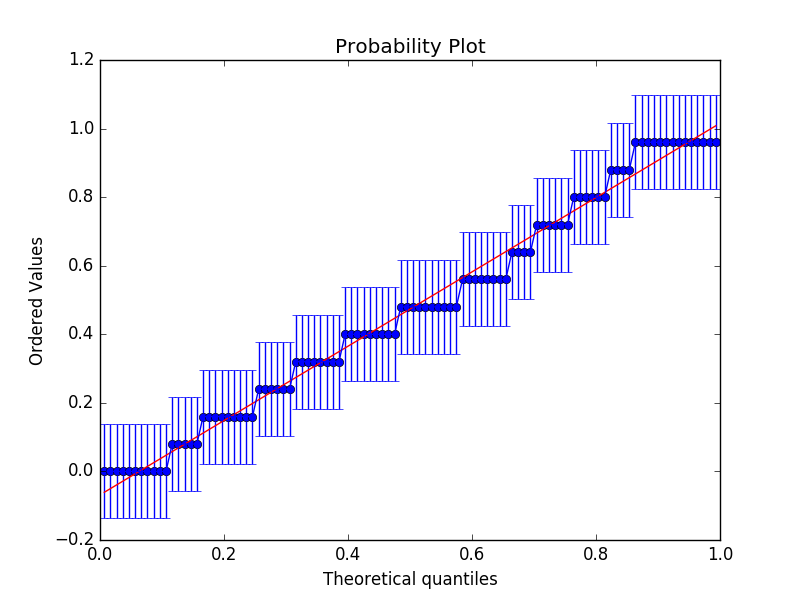}
  \caption{P-P plot for $W = 23, M = 92, K = 10, T = 0.137$}\label{fig:9223}
\endminipage
\end{figure}

\subsection*{Experimental Setup for Section 4.2 Use Case}
\textbf{Temporal Grouping.} We consider articles from the last four years of the corpus (2004-2007), dividing this time frame into four time periods, one per year, taking articles (from Foreign desk only for Table 
\ref{table: grouppvals}, from all desks for Table \ref{table: grouppvalsall}) in a day to form a document. For the preliminary results presented in this paper we sampled the first month (January) in each of the four time periods, with $X_1$ being 2004, $X_2$ being 2005 and so forth.

\textbf{Desk Grouping.} We select four desks - foreign, business, arts and culture, and national, taking articles (from year 2005 only for Table \ref{table: grouppvals}, from all years for Table \ref{table: grouppvalsall}), again taking all articles in a day to form a document. All articles in the first desk are in $X_1$; all articles in the second desk are in $X_2$ and so forth. 

\textbf{Evaluation.} We used $k=5$ topics, and replicated each experiment ten times, averaging the p-values over ten trials. We also report the number of trials (out of ten) the p-value is less than the rejection boundary. 

\subsection*{Table of Results for All Years and All Desks}
\begin{table}[h!]
\centering
\begin{tabular}{| c c c c|}
\hline
Matrix &Scope & p-value $\rho$ & Rejects\\
\hline
$X$ & &  0.0 & 10 \\
\hdashline
$X_1$ & 2004 & 0.15 & 6 \\
$X_2$ & 2005 & 0.0 & 10\\
$X_3$ & 2006 & 0.09 & 7\\
$X_4$ & 2007 & 0.04 & 9\\
\hline
\end{tabular}
\begin{tabular}{| c c c c|}
\hline
Matrix & Scope & p-value $\rho$ & Rejects\\
\hline
$X$ & & 0.04 & 9 \\
\hdashline
$X_1$ & Foreign & 0.648 & 0 \\
$X_2$ & Business & 0.728 & 1 \\
$X_3$ & Arts & 0.624 & 0\\
$X_4$ & National & 0.0 & 10\\
\hline
\end{tabular}
\caption{\textbf{Left}: Testing for temporal structure in New York Times articles from \textbf{all desks}: p-values for proposed parametric bootstrap test on entire document-term matrix vs. same matrix broken into four sub-matrices by time.  \\
\textbf{Right}: Testing for desk structure in New York Times articles from \textbf{all years}: p-values for proposed parametric bootstrap test on entire document-term matrix vs. same matrix broken into four sub-matrices by desk.
}
\label{table: grouppvalsall}
\end{table}


\end{document}